%% file: main.tex
\def\year{2020}\relax
\documentclass[letterpaper]{article} 
\usepackage{aaai20}  
\usepackage{times}  
\usepackage{helvet} 
\usepackage{courier}  
\usepackage[hyphens]{url}  
\usepackage{graphicx} 
\usepackage{graphicx}  
\usepackage{subcaption}
\usepackage{mathtools}
\usepackage{amsmath}
\usepackage{amssymb}

\DeclareMathAlphabet{\pazocal}{OMS}{zplm}{m}{n}

\newcommand{\citet}[1]{\citeauthor{#1} \shortcite{#1}}
\newcommand{\citep}{\cite}

\title{Assessing the Adversarial Robustness of Monte Carlo and\\ Distillation Methods for Deep Bayesian Neural Network Classification}
\author{ \Large \textbf{Meet P. Vadera\textsuperscript{\rm 1, *}, Satya Narayan Shukla\textsuperscript{\rm 1, *}, Brian Jalaian\textsuperscript{\rm 2}, and Benjamin M. Marlin\textsuperscript{\rm 1}}\\ 
\textsuperscript{\rm 1}University of Massachusetts Amherst, \textsuperscript{\rm 2}US Army Research Laboratory\\ 
\{mvadera,snshukla,marlin\}@cs.umass.edu, brian.jalaian.ctr@mail.mil	 
}

\begin{document}

\maketitle

\let\thefootnote\relax\footnotetext{\textsuperscript{\rm *} Equal contribution. Presented at SafeAI 2020.}

\input{abstract.tex}

\input{introduction.tex}

\input{related_work.tex}

\input{methods.tex}

\input{experiments.tex}

\input{discussion.tex}

\section*{Acknowledgments}
This work was partially supported by the US Army Research Laboratory under cooperative agreement W911NF-17-2-0196. The   views   and   conclusions  contained  in  this  document  are  those  of  the  authors  and  should  not  be interpreted as representing the official policies,  either  expressed  or  implied,  of  the  Army  Research  Laboratory  or  the  US  government.

\bibliographystyle{aaai}
\bibliography{references}
\end{document}

%% file: abstract.tex
\begin{abstract}
In this paper, we consider the problem of assessing the adversarial robustness of deep neural network models under both Markov chain Monte Carlo (MCMC) and Bayesian Dark Knowledge (BDK) inference approximations. We characterize the robustness of each method to two types of adversarial attacks: the fast gradient sign method (FGSM) and projected gradient descent (PGD). We show that full MCMC-based inference has excellent robustness, significantly outperforming standard point estimation-based learning. On the other hand, BDK provides marginal improvements. As an additional contribution, we present a storage-efficient approach to computing adversarial examples for large Monte Carlo ensembles using both the FGSM and PGD attacks.
\end{abstract}

%% file: introduction.tex
\section{Introduction}
Deep learning models have shown promising results in areas including computer vision, natural language processing, speech recognition, and more \citep{krizhevsky2012imagenet,graves2013hybrid,graves2013speech,Huang2016DenselyCC,Devlin2018BERTPO}. 
Despite these advances, deep neural networks are well-known to be vulnerable to adversarial examples. An adversarial example is an example that differs from a natural example through a low-norm additive perturbation while resulting in an incorrect prediction relative to the unperturbed example \cite{goodfellow2014explaining}. The lack of robustness to adversarial examples is a crucial barrier to the safe deployment of deep learning models in many applications.  



One potential source of adversarial examples in traditional point-estimated deep neural network models derives from the fact that the decision boundary geometry can be minimally constrained away from the training data during learning. This can lead to quite arbitrary decision boundaries away from the training data, which may be easily attackable \cite{nguyen2015}. By contrast, Bayesian inference methods result in predictable and well-behaved decision boundary geometry off of the training data due to the Bayesian model averaging effect when computing the posterior predictive distribution. If many decision boundary geometries are all equally likely in a given region of feature space, the posterior predictive distribution will average over all of them resulting in both posterior class probabilities and decision boundary geometry with improved smoothness, which may be harder to attack with low-norm adversarial perturbations. 

Indeed, \citet{gal2018sufficient} present theoretical evidence that under certain sufficient conditions there exists no adversarial examples for a Bayesian classification model. However, \citet{liu2018advbnn} show that Bayesian neural networks implemented using variational inference lack robustness against adversarial samples, and subsequently present a method that attempts to make them more robust. 


Markov Chain Monte Carlo (MCMC) methods are one of the primary alternatives to variational methods for performing approximate inference in Bayesian neural networks. While we defer the background on Bayesian neural networks to the next section, it is important to note that MCMC methods give an unbiased estimate to the parameter posterior and the posterior predictive distribution, while variational methods are typically biased. However, MCMC methods require materializing or storing samples of the parameter posterior in order to make predictions, which can have high computational complexity and storage cost.

To help overcome these problems, \citet{balan2015bayesian} introduced a model distillation method referred to as \emph{Bayesian Dark Knowledge} (BDK). In the classification setting, Bayesian Dark Knowledge attempts to compress the Bayesian posterior predictive distribution induced by the full parameter posterior of a ``teacher" network into a single, compact ``student" network. 
The major advantage of this approach is that the computational complexity of prediction at test time is drastically reduced. This method has been shown to successfully reproduce the full posterior predictive distribution on test data drawn from the training distribution \cite{balan2015bayesian}.


In this paper, we consider the problem of assessing the adversarial robustness of deep neural network models under both the MCMC and Bayesian Dark Knowledge approximations. We characterize the robustness of each method to two types of adversarial attacks: the fast gradient sign method (FGSM) \cite{goodfellow2014explaining} and projected gradient descent (PGD) \cite{madry_pgd}. We consider the case of a basic convolutional neural network (CNN) architecture and the MNIST and CIFAR10 data sets. Interestingly, we show that full MCMC-based inference has excellent robustness to these adversarial attacks, significantly outperforming standard point estimation-based learning, while BDK only provides marginal improvements. This indicates that the BDK distillation procedure is failing to fully capture the structure of the true posterior predictive distribution off of the training data. As an additional contribution, we present a storage-efficient approach to computing adversarial examples for large Monte Carlo ensembles using both the FGSM and PGD attacks.

%% file: related_work.tex
\section{Background}
\textbf{Bayesian Neural Networks:} Let $p(y|\mathbf{x}, \theta)$ represent the probability distribution induced by a deep neural network classifier over classes $y\in\pazocal{Y}=\{1,..,C\}$ given feature vectors $\mathbf{x}\in\mathbb{R}^D$. The most common way to fit a model of this type given a data set $\pazocal{D} =\{(\mathbf{x}_i,y_i)|1\leq i\leq N\}$ is to use maximum conditional likelihood estimation, or equivalently, cross entropy loss minimization (or their penalized or regularized variants). However, when the volume of labeled data is low, there can be multiple advantages to considering a full Bayesian treatment of the model. Instead of attempting to find the single (locally) optimal parameter set $\theta_*$ according to a given criterion, Bayesian inference uses Bayes rule to define the posterior distribution $p(\theta|\pazocal{D},\theta^0)$ over the unknown parameters $\theta$ given a prior distribution $P(\theta|\theta^0)$ with prior parameters $\theta^0$ as seen in Equation \ref{eq:posterior}. 
\begin{align}
\allowdisplaybreaks[4]
\label{eq:posterior}
    p(\theta|\pazocal{D},\theta^0)& = \frac{p(\pazocal{D}|\theta)p(\theta|\theta^0)}{\int p(\pazocal{D}|\theta)p(\theta|\theta^0) d\theta}\\
\label{eq:posterior_predictive}
p(y| \mathbf{x}, \pazocal{D},\theta^0) &= \int p(y|\mathbf{x}, \theta) p(\theta|\pazocal{D},\theta^0) d\theta
\end{align}
For prediction problems in machine learning, the quantity of interest is typically not the parameter posterior itself, but the posterior predictive distribution $p(y| \mathbf{x}, \pazocal{D},\theta^0)$ obtained from it as seen in  Equation  \ref{eq:posterior_predictive}. The primary problem with applying Bayesian inference to neural network models is that the distributions $p(\theta|\pazocal{D},\theta^0)$ and $p(y| \mathbf{x}, \pazocal{D},\theta^0)$ are not available in closed form, so approximations are required.

Most Bayesian inference approximations studied in the machine learning literature are based on variational inference (VI) \citep{jordan1999introduction} or Markov Chain Monte Carlo (MCMC) methods \citep{Neal:1996:BLN:525544,welling2011bayesian}.
In VI, an auxiliary distribution $q_{\phi} (\theta)$  is defined to approximate the true parameter posterior $p(\theta | \pazocal{D},\theta^0)$. 
%
%
The main drawback of VI and related methods is that they  typically result in biased posterior estimates for complex posterior distributions. MCMC methods provide an alternative family of sampling-based posterior approximations that are unbiased. The samples generated using MCMC methods can then be used to approximate the posterior predictive distribution using a Monte Carlo average as shown in Equation \ref{eq:MC_posterior_predictive}.
\begin{align}
\allowdisplaybreaks[4]
\label{eq:MC_posterior_predictive}
p(y| \mathbf{x}, \pazocal{D},\theta^0) 
&\approx  \frac{1}{T}\sum_{t=1}^T p(y|\mathbf{x}, \theta_t); \;\;\;\;
\theta_t \sim p(\theta|\pazocal{D},\theta^0)
\end{align}
\citet{Neal:1996:BLN:525544} first addressed the problem of Bayesian inference in neural networks using Hamiltonian Monte Carlo (HMC) to provide a set of posterior samples. 
The stochastic gradient Langevin dynamics (SGLD) sampling method improves on HMC by enabling sampling based on minibatches of data to improve the time computational complexity of sampling  \citep{welling2011bayesian}; however, the problem of needing to compute over a large set of samples when making predictions at test or deployment time still remains. Bayesian Dark Knowledge \citep{balan2015bayesian} aims at reducing the test-time computational complexity of Monte Carlo-based approximations for neural networks by distilling the posterior predictive distribution (approximated by Equation \ref{eq:MC_posterior_predictive}) of a neural network into another neural network. We will discuss the details of both methods in Section \ref{sec:methods}.

\textbf{Adversarial Attacks:} Adversarial examples are carefully crafted perturbations to a classifier input that are designed to mislead a classifier while being as imperceptible as possible. Based on the information available to the adversary, these attacks are broadly categorized as {\it white-box} and {\it black-box} attacks. Most methods for adversarial attacks on deep learning models operate in the white-box setting \citep{goodfellow2014explaining,madry_pgd,BIM,deepfool,features,CarliniW16a}, where the model being attacked, and its gradients, are assumed to be fully known. Conversely, the black-box setting \citep{brendel,cheng2018query,zoo,autozoom,nes} requires an attacker to find an adversarial perturbation when its only access to  the model is via labeling queries.

The most successful adversarial attacks use gradient-based
optimization methods. For example, the Fast Gradient Sign Method (FGSM) \citep{goodfellow2014explaining} is a one-step method that uses the sign of the gradient to create adversarial examples:
\begin{equation}
\mathbf{x}_{adv} = \mathbf{x} + \epsilon \; \text{sign}(\nabla_{\mathbf{x}} \mathcal{L}(\theta, \mathbf{x}, y))  , 
\end{equation}  
where $\mathcal{L}$ is the standard cross-entropy loss computed using the model prediction given input $\mathbf{x}$ and the original label $y$, and $\epsilon$ governs the magnitude of the perturbation introduced and can be thought of as a step size. \citet{kurakin2017} extended this to a multi-step variant which is more powerful than single step FGSM. Both the methods are, however, limited to generating $\ell_\infty$-bounded perturbations. \citet{madry_pgd} introduced a more general multi-step variant, which is essentially projected gradient descent (PGD) on the negative loss function. FGSM-based attacks can be seen as specific instances of PGD under $\ell_\infty$-bounded perturbations. 
The main emphasis of the PGD attack is to apply FGSM $k$ times (number of iterations) with step-size $\alpha \leq \epsilon/k$, where $\epsilon$ is the maximum distortion (i.e., attack strength) of the adversarial example compared to the original input. The resulting adversarial example $\mathbf{x}^{t+1}$ corresponding to input $x$ is computed as follows:
\begin{equation}
\mathbf{x}^{t+1} = \Pi_{\mathbf{x}+S}(\mathbf{x}^t + \alpha\; \text{sign}(\nabla_{\mathbf{x}} \mathcal{L}(\theta, \mathbf{x}, y))) 
\end{equation} 
where $\Pi_{x+S}$ is the projection onto the $\ell_\infty$ ball of radius $\epsilon$ centred at $\mathbf{x}$.

%% file: methods.tex
\section{Methods}
\label{sec:methods}
\subsection{Approximate Bayesian Inference Methods }
For implementing approximate Bayesian inference in neural networks, we adopt the use of the stochastic gradient Langevin dynamics (SGLD) \citep{welling2011bayesian} method as a computationally efficient MCMC sampler. We compare the adversarial robustness of the SGLD approximation to the posterior predictive distribution to that provided by Bayesian dark knowledge (BDK) \cite{balan2015bayesian}, which is drastically more efficient to deploy. SGLD is also used at the core of the BDK posterior distillation algorithm. We begin by reviewing SGLD, and then describe BDK. 

Let $p(\theta| \lambda)$ be the prior distribution over the network model parameters $\theta$. The prior distribution is selected to be a spherical Gaussian distribution centered at $0$ with precision denoted by $\lambda$. We define $\pazocal{S}$ to be a minibatch of size M drawn from $\pazocal{D}$. Let the total number of training samples in $\pazocal{D}$ be denoted by $N$. $\theta_t$ denotes the parameter set sampled from the model at sampling iteration $t$, while $\eta_t$ denotes the learning rate at iteration $t$. The Langevin noise is denoted by $z_t \sim \pazocal{N}(0, \eta_t I )$.  The sampling update for SGLD can be written as seen below where $p\left(y_{i} | x_{i}, \theta_{t}\right)$ is the likelihood:
{\footnotesize
\begin{equation}
\Delta \theta_{t+1} \!=\!\frac{\eta_{t}}{2}\!\left(\!\nabla_{\theta} \log p(\theta | \lambda)\!+\!\frac{N}{M} \sum_{i \in \pazocal{S}} \nabla_{\theta} \log p\left(y_{i} | x_{i}, \theta_{t}\right)\!\right)+ z_{t}.
\end{equation}
}
By running this update, we produce a sequence of samples that converges to the posterior distribution of the model we are sampling from. This set of samples forms a Monte Carlo ensemble and is used in Equation \ref{eq:MC_posterior_predictive} when making predictions. In BDK terminology, the model we sample from is referred to as the ``teacher" model and the set of sampled models is referred to as the ``teacher ensemble." BDK aims to compress the true posterior predictive distribution as approximated by a teacher ensemble into a compact, feed-forward neural network model (referred to as the "student" model) using distillation methods to avoid the need to store samples and compute predictions using Equation \ref{eq:MC_posterior_predictive} at deployment time. 

To learn the student model, BDK generates a batch of samples $\pazocal{S'}$. $\pazocal{S'}$ is obtained by adding Gaussian noise of small magnitude to $\pazocal{S}$. The output of the teacher model on samples $\{(\mathbf{x'}, y')\} \in \pazocal{S'}$ is approximated using the current sample from the teacher model: $p(y'|\mathbf{x'}, \theta_{t+1})$. Similarly, the output of the student model, parameterized by $\omega_t$, is computed as $p(y'|\mathbf{x'}, \omega_t)$. The objective function for learning the student is the KL divergence between the teacher model's predictive distribution and the student model's predictive distribution: $\pazocal{L}( \omega_t| \pazocal{S}', \theta_{t+1}) = \sum_{(\mathbf{x'}, y') \in \pazocal{S'} }\textrm{KL}(p(y'|\mathbf{x'}, \theta_{t+1}) || p(y'|\mathbf{x'}, \omega_t))$ and we run a single optimization iteration on it to obtain $\omega_{t+1}$. This process of sequentially computing $\theta_{t+1}$ and $\omega_{t+1}$ is repeated until convergence. 

\subsection{Attacking Approximate Bayesian Inference}
To investigate the adversarial robustness of the Monte Carlo ensembles produced produced by SGLD and the distilled student models produced by BDK, we apply two common forms of attacks adopted in the literature: PGD and FGSM. Both of these attacks rely on computing the gradient of the cross-entropy loss function between the model output and labels w.r.t the inputs. This presents a computational challenge when the model we must attack is a large Monte Carlo ensemble consisting of hundreds or thousands of models. Indeed, even explicitly storing all of the models in the ensemble can be a challenge. To address this problem, we leverage the fact that the gradient of the loss w.r.t input of the ensemble is a sum of the gradients of the loss w.r.t the input for each model in the ensemble. For an ensemble $\{\theta_i\}_{i=1}^K$ consisting of $K$ models sampled from the posterior, the gradient of the loss $\pazocal{L}(\cdot; \theta_{1:K})$ w.r.t. input $\mathbf{x}$ can be expressed as:
\begin{equation}
\nabla_\mathbf{x} \pazocal{L}(y, \mathbf{x}; \theta_{1:K}) = \frac{1}{K} \sum_{i=1}^K \nabla_\mathbf{x} \pazocal{L}(\cdot ; \theta_i) 
\end{equation}
The previous equation follows directly from Equation \ref{eq:MC_posterior_predictive}. As can be seen clearly, we only need access to a single model from the ensemble at a time and accumulate the gradients to obtain the gradient of the ensemble. This enables us to perform FGSM and PGD attacks with constant memory while scaling up the number of models in the ensemble. Further, if we have access to the data, we only actually need to store a starting model and a random seed. The rest of the elements of the ensemble can be sequentially materialized by re-running the SGLD sampling iteration. This allows us to also generate attacks for large ensembles while using constant storage cost. 

We note that this corresponds to an incredibly strong attack against a Bayesian model as we effectively assume that we have access to every element of the approximating Monte Carlo ensemble, despite the fact that the samples could be updated at any time. Finally, we note that attacking the BDK student model is straightforward since the student is a standard feed-forward model. This attack requires no additional modifications to the original algorithms. 

%% file: experiments.tex
\section{Experiments}

\begin{figure*}[t]
\centering
\begin{subfigure}[b]{0.75\textwidth}
   \includegraphics[width=1\linewidth]{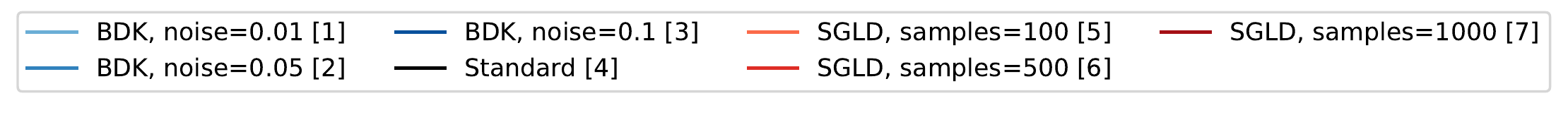}
\end{subfigure}

\begin{subfigure}[b]{0.31\textwidth}
   \includegraphics[width=1\linewidth]{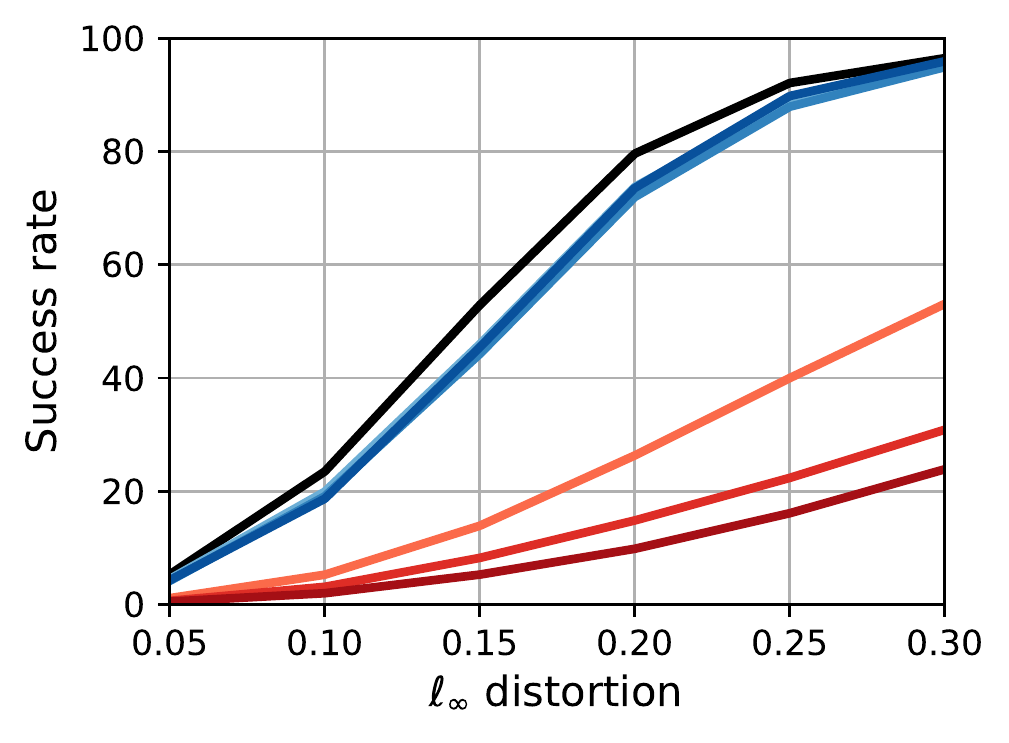}
   \caption{FGSM Success rate}
   \label{fig:mnist_fgsm_success} 
\end{subfigure}
\begin{subfigure}[b]{0.31\textwidth}
  \includegraphics[width=1\linewidth]{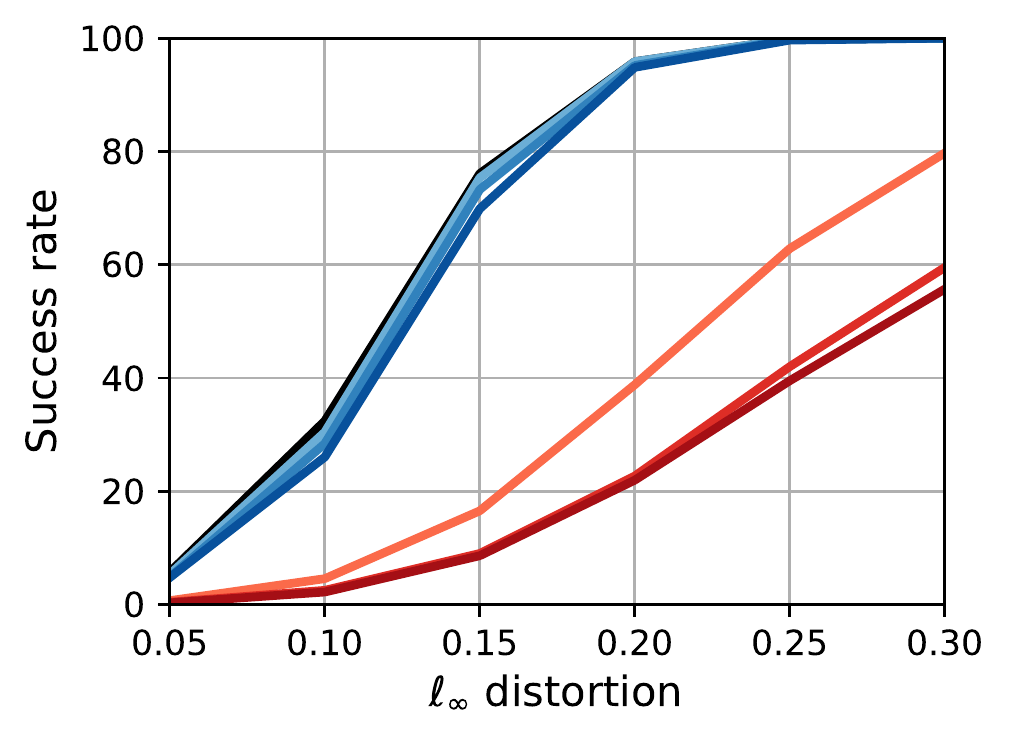}
  \caption{PGD Success rate}
  \label{fig:mnist_pgd_success}
\end{subfigure}
\begin{subfigure}[b]{0.31\textwidth}
  \includegraphics[width=1\linewidth]{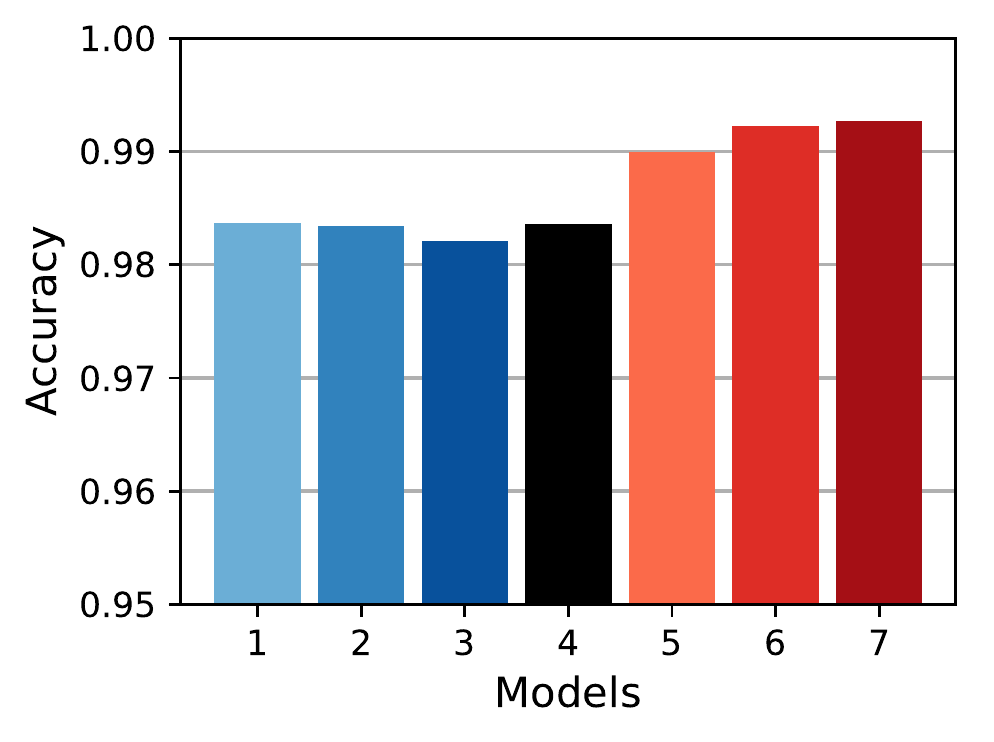}
  \caption{Accuracy}
  \label{fig:mnist_acc}
\end{subfigure}
%
\centering
\begin{subfigure}[d]{0.31\textwidth}
   \includegraphics[width=1\linewidth]{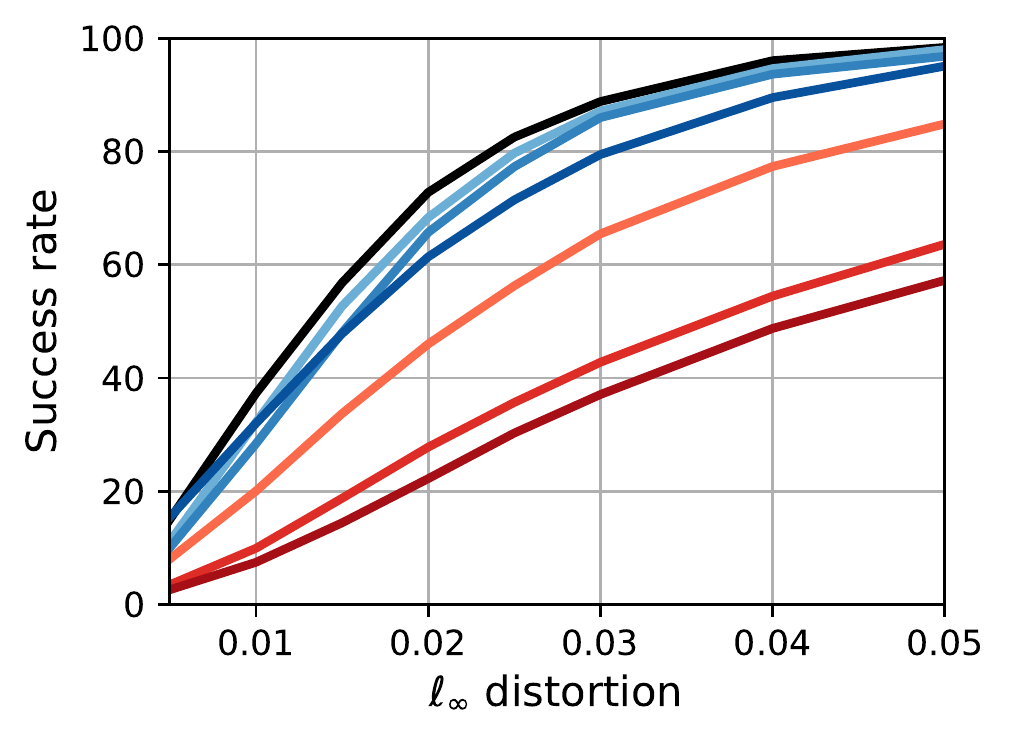}
   \caption{FGSM Success rate}
   \label{fig:cifar10_fgsm_success} 
\end{subfigure}
\begin{subfigure}[e]{0.31\textwidth}
  \includegraphics[width=1\linewidth]{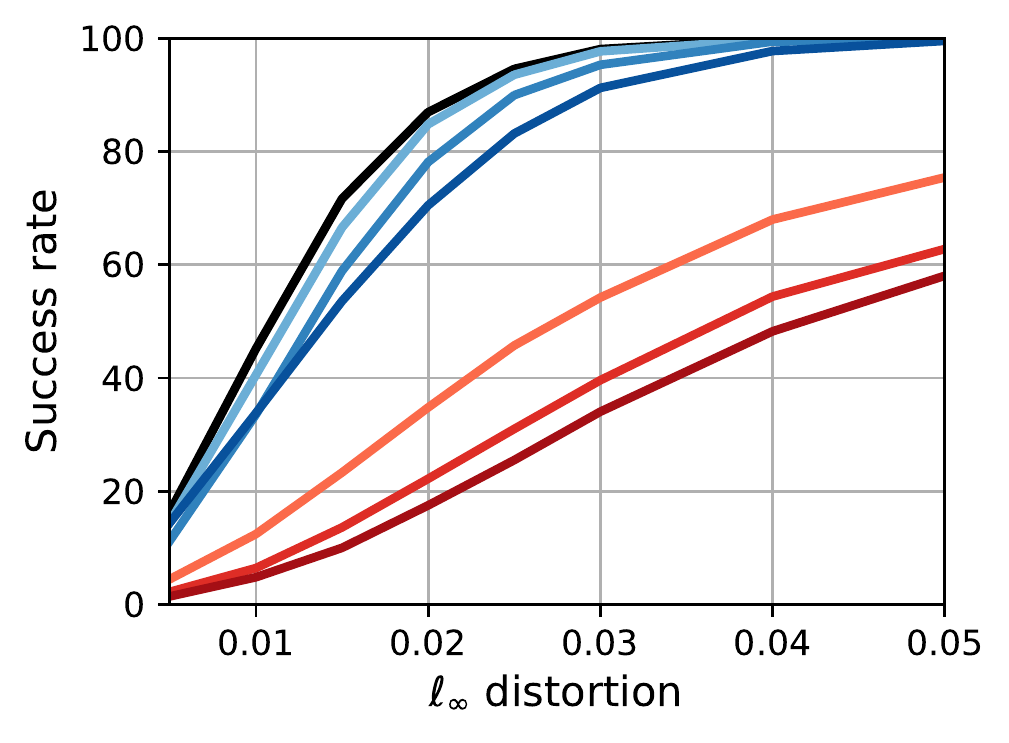}
  \caption{PGD Success rate}
  \label{fig:cifar10_pgd_success}
\end{subfigure}
\begin{subfigure}[f]{0.31\textwidth}
  \includegraphics[width=1\linewidth]{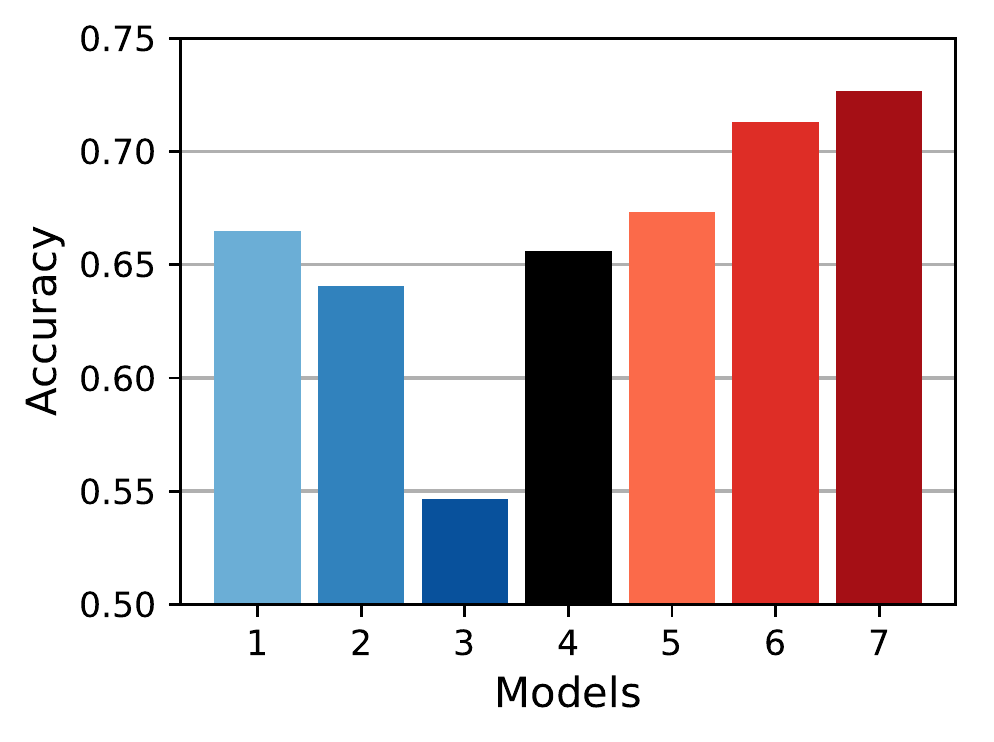}
  \caption{Accuracy}
  \label{fig:cifar10_acc}
\end{subfigure}
\caption{Performance comparison for untargeted $\ell_\infty$ attacks on MNIST (top row) and CIFAR10 (bottom row).}
\label{fig:mnist}
\label{fig:cifar10}
\end{figure*}

Our experiments focus on the untargeted attack setting where the goal is to cause the model to misclassify inputs that are otherwise correctly classified. We consider the $\ell_\infty$ threat model for the experiments. We study the adversarial robustness of SGLD Monte Carlo ensembles and distilled BDK student models. We compare their performance to standard point estimation of the same model (we use Adam as the optimizer). We consider the white-box FGSM \citep{goodfellow2014explaining} and PGD \citep{madry_pgd} attacks. We evaluate the models on MNIST \citep{mnist} and CIFAR-10 \citep{cifar10} datasets in terms of attack success rate and accuracy on unperturbed test data.

We define the attack success rate as the percentage of the number of inputs successfully perturbed using the attack method to the total number of input images that we attempt to perturb. In all experiments, images that are already misclassified by any model are excluded from the attack set.  All results are based on the test set, which consists of 10000 examples for each data set.

\textbf{Models}: We utilize CNNs for both the teacher ensemble and the student model. Further, for a given data set, we use the same architecture for the teacher as well as the student. For MNIST, we use the following architecture: Input(1, (28,28))-Conv(num\_kernels=10, kernel\_size=4, stride=1) - MaxPool(kernel\_size=2) - Conv(num\_kernels=20, kernel\_size=4, stride=1) - MaxPool(kernel\_size=2) - FC (80) - FC (output). For CIFAR10, we utilize the following architecture: Input(3, (32,32)) - Conv(num\_kernels=16, kernel\_size=5) - MaxPool(kernel\_size=2) - Conv(num\_kernels=32, kernel\_size=5) - MaxPool(kernel\_size=2) - FC(200) - FC (50) - FC (output).

\textbf{Model and Distillation Hyperparameters}: We run the SGLD and distillation procedure using the following hyperparameters: fixed teacher learning rate $\eta_t = 4 \times 10^{-6}$ for MNIST and $\eta_t = 3 \times 10^{-6}$ for CIFAR10, teacher prior precision $\lambda = 10$, initial student learning rate $\rho_t = 10^{-3}$, burn-in iterations $B=1000$ for MNIST and $B=10000$ for CIFAR10, thinning interval $\tau = 100$, and total training iterations $T= 10^6$. For training the student model, we use the \emph{Adam} algorithm and set a learning schedule for the student such that it halves its learning rate every 200 epochs for MNIST, and every 400 epochs for CIFAR10.

\textbf{Results:} Figure \ref{fig:mnist} compares the performance of standard, SGLD and BDK models on MNIST and CIFAR10. We vary the $\ell_\infty$ distortion from $0.05$ to $0.3$ for MNIST and from $0.005$ to $0.05$ for CIFAR10 and report the attack success rates of FGSM and PGD. For PGD attacks, we run 40 iterations with step size of  0.05 for MNIST and 0.005 for CIFAR10. Note that lower attack success rate implies higher adversarial robustness. We can see that the SGLD Monte Carlo ensembles are significantly more robust to the adversarial attack than the standard models. We also observe that the adversarial robustness as well as accuracy of SGLD models increase as the number of models in the ensemble increases. We see that the BDK student models only provide a marginal increase in robustness as the amount of noise used for distillation increases. However, this comes at the price of accuracy as shown in Figures \ref{fig:mnist_acc} and \ref{fig:cifar10_acc} where we observe that test accuracy on unperturbed test data decreases with increasing noise for BDK student models.

%% file: discussion.tex
\section{Conclusions and Future Directions}
We have considered the problem of assessing the adversarial robustness of deep neural network models under both the Markov Chain Monte Carlo (MCMC) and Bayesian Dark Knowledge (BDK) inference approximations. Interestingly, our results show that full MCMC-based inference has excellent robustness, significantly outperforming standard point estimation-based learning, while BDK provides marginal improvement. A key direction for future work will thus be to further investigate the failure of BDK to fully capture the posterior predictive distribution to see if its deployment-time computational advantages over MCMC-based methods can be preserved while enhancing its adversarial robustness. 

%% file: main.bbl
\begin{thebibliography}{}

\bibitem[\protect\citeauthoryear{Balan \bgroup et al\mbox.\egroup
  }{2015}]{balan2015bayesian}
Balan, A.~K.; Rathod, V.; Murphy, K.~P.; and Welling, M.
\newblock 2015.
\newblock Bayesian dark knowledge.
\newblock In {\em Advances in Neural Information Processing Systems},
  3438--3446.

\bibitem[\protect\citeauthoryear{Brendel, Rauber, and Bethge}{2017}]{brendel}
Brendel, W.; Rauber, J.; and Bethge, M.
\newblock 2017.
\newblock Decision-based adversarial attacks: Reliable attacks against
  black-box machine learning models.
\newblock {\em arXiv preprint arXiv:1712.04248}.

\bibitem[\protect\citeauthoryear{Carlini and Wagner}{2016}]{CarliniW16a}
Carlini, N., and Wagner, D.~A.
\newblock 2016.
\newblock Towards evaluating the robustness of neural networks.
\newblock {\em CoRR} abs/1608.04644.

\bibitem[\protect\citeauthoryear{Chen \bgroup et al\mbox.\egroup }{2017}]{zoo}
Chen, P.-Y.; Zhang, H.; Sharma, Y.; Yi, J.; and Hsieh, C.-J.
\newblock 2017.
\newblock Zoo: Zeroth order optimization based black-box attacks to deep neural
  networks without training substitute models.
\newblock In {\em Proceedings of the 10th ACM Workshop on Artificial
  Intelligence and Security},  15--26.
\newblock ACM.

\bibitem[\protect\citeauthoryear{Cheng \bgroup et al\mbox.\egroup
  }{2018}]{cheng2018query}
Cheng, M.; Le, T.; Chen, P.-Y.; Yi, J.; Zhang, H.; and Hsieh, C.-J.
\newblock 2018.
\newblock Query-efficient hard-label black-box attack: An optimization-based
  approach.
\newblock {\em arXiv preprint arXiv:1807.04457}.

\bibitem[\protect\citeauthoryear{Devlin \bgroup et al\mbox.\egroup
  }{2018}]{Devlin2018BERTPO}
Devlin, J.; Chang, M.-W.; Lee, K.; and Toutanova, K.
\newblock 2018.
\newblock Bert: Pre-training of deep bidirectional transformers for language
  understanding.
\newblock In {\em NAACL-HLT}.

\bibitem[\protect\citeauthoryear{Gal and Smith}{2018}]{gal2018sufficient}
Gal, Y., and Smith, L.
\newblock 2018.
\newblock Sufficient conditions for idealised models to have no adversarial
  examples: a theoretical and empirical study with bayesian neural networks.
\newblock {\em arXiv preprint arXiv:1806.00667}.

\bibitem[\protect\citeauthoryear{Goodfellow, Shlens, and
  Szegedy}{2014}]{goodfellow2014explaining}
Goodfellow, I.~J.; Shlens, J.; and Szegedy, C.
\newblock 2014.
\newblock Explaining and harnessing adversarial examples.
\newblock {\em arXiv preprint arXiv:1412.6572}.

\bibitem[\protect\citeauthoryear{Graves, Jaitly, and
  Mohamed}{2013}]{graves2013hybrid}
Graves, A.; Jaitly, N.; and Mohamed, A.-r.
\newblock 2013.
\newblock Hybrid speech recognition with deep bidirectional lstm.
\newblock In {\em Automatic Speech Recognition and Understanding (ASRU), 2013
  IEEE Workshop on},  273--278.
\newblock IEEE.

\bibitem[\protect\citeauthoryear{Graves, Mohamed, and
  Hinton}{2013}]{graves2013speech}
Graves, A.; Mohamed, A.-r.; and Hinton, G.
\newblock 2013.
\newblock Speech recognition with deep recurrent neural networks.
\newblock In {\em Acoustics, speech and signal processing (icassp), 2013 ieee
  international conference on},  6645--6649.
\newblock IEEE.

\bibitem[\protect\citeauthoryear{Huang \bgroup et al\mbox.\egroup
  }{2016}]{Huang2016DenselyCC}
Huang, G.; Liu, Z.; van~der Maaten, L.; and Weinberger, K.~Q.
\newblock 2016.
\newblock Densely connected convolutional networks.
\newblock {\em 2017 IEEE Conference on Computer Vision and Pattern Recognition
  (CVPR)}  2261--2269.

\bibitem[\protect\citeauthoryear{Ilyas \bgroup et al\mbox.\egroup }{2018}]{nes}
Ilyas, A.; Engstrom, L.; Athalye, A.; and Lin, J.
\newblock 2018.
\newblock Black-box adversarial attacks with limited queries and information.
\newblock {\em arXiv preprint arXiv:1804.08598}.

\bibitem[\protect\citeauthoryear{Jordan \bgroup et al\mbox.\egroup
  }{1999}]{jordan1999introduction}
Jordan, M.~I.; Ghahramani, Z.; Jaakkola, T.~S.; and Saul, L.~K.
\newblock 1999.
\newblock An introduction to variational methods for graphical models.
\newblock {\em Machine learning} 37(2):183--233.

\bibitem[\protect\citeauthoryear{Krizhevsky, Hinton, and
  others}{2009}]{cifar10}
Krizhevsky, A.; Hinton, G.; et~al.
\newblock 2009.
\newblock Learning multiple layers of features from tiny images.
\newblock Technical report, Citeseer.

\bibitem[\protect\citeauthoryear{Krizhevsky, Sutskever, and
  Hinton}{2012}]{krizhevsky2012imagenet}
Krizhevsky, A.; Sutskever, I.; and Hinton, G.~E.
\newblock 2012.
\newblock Imagenet classification with deep convolutional neural networks.
\newblock In {\em Advances in neural information processing systems},
  1097--1105.

\bibitem[\protect\citeauthoryear{Kurakin, Goodfellow, and Bengio}{2016}]{BIM}
Kurakin, A.; Goodfellow, I.; and Bengio, S.
\newblock 2016.
\newblock Adversarial examples in the physical world.
\newblock {\em arXiv preprint arXiv:1607.02533}.

\bibitem[\protect\citeauthoryear{Kurakin, Goodfellow, and
  Bengio}{2017}]{kurakin2017}
Kurakin, A.; Goodfellow, I.~J.; and Bengio, S.
\newblock 2017.
\newblock Adversarial machine learning at scale.

\bibitem[\protect\citeauthoryear{Lecun \bgroup et al\mbox.\egroup
  }{1998}]{mnist}
Lecun, Y.; Bottou, L.; Bengio, Y.; and Haffner, P.
\newblock 1998.
\newblock Gradient-based learning applied to document recognition.
\newblock In {\em Proceedings of the IEEE},  2278--2324.

\bibitem[\protect\citeauthoryear{Liu \bgroup et al\mbox.\egroup
  }{2019}]{liu2018advbnn}
Liu, X.; Li, Y.; Wu, C.; and Hsieh, C.-J.
\newblock 2019.
\newblock Adv-{BNN}: Improved adversarial defense through robust bayesian
  neural network.
\newblock In {\em International Conference on Learning Representations}.

\bibitem[\protect\citeauthoryear{Madry \bgroup et al\mbox.\egroup
  }{2017}]{madry_pgd}
Madry, A.; Makelov, A.; Schmidt, L.; Tsipras, D.; and Vladu, A.
\newblock 2017.
\newblock Towards deep learning models resistant to adversarial attacks.
\newblock {\em arXiv preprint arXiv:1706.06083}.

\bibitem[\protect\citeauthoryear{Moosavi-Dezfooli, Fawzi, and
  Frossard}{2016}]{deepfool}
Moosavi-Dezfooli, S.-M.; Fawzi, A.; and Frossard, P.
\newblock 2016.
\newblock Deepfool: a simple and accurate method to fool deep neural networks.
\newblock In {\em Proceedings of the IEEE conference on computer vision and
  pattern recognition},  2574--2582.

\bibitem[\protect\citeauthoryear{Neal}{1996}]{Neal:1996:BLN:525544}
Neal, R.~M.
\newblock 1996.
\newblock {\em Bayesian Learning for Neural Networks}.
\newblock Berlin, Heidelberg: Springer-Verlag.

\bibitem[\protect\citeauthoryear{Nguyen, Yosinski, and
  Clune}{2015}]{nguyen2015}
Nguyen, A.~M.; Yosinski, J.; and Clune, J.
\newblock 2015.
\newblock Deep neural networks are easily fooled: High confidence predictions
  for unrecognizable images.
\newblock In {\em CVPR},  427--436.
\newblock IEEE Computer Society.

\bibitem[\protect\citeauthoryear{Sabour \bgroup et al\mbox.\egroup
  }{2015}]{features}
Sabour, S.; Cao, Y.; Faghri, F.; and Fleet, D.~J.
\newblock 2015.
\newblock Adversarial manipulation of deep representations.
\newblock {\em arXiv preprint arXiv:1511.05122}.

\bibitem[\protect\citeauthoryear{Tu \bgroup et al\mbox.\egroup
  }{2019}]{autozoom}
Tu, C.; Ting, P.; Chen, P.; Liu, S.; Zhang, H.; Yi, J.; Hsieh, C.; and Cheng,
  S.
\newblock 2019.
\newblock Autozoom: Autoencoder-based zeroth order optimization method for
  attacking black-box neural networks.
\newblock In {\em The Thirty-Third {AAAI} Conference on Artificial
  Intelligence, {AAAI} 2019, The Thirty-First Innovative Applications of
  Artificial Intelligence Conference, {IAAI} 2019, The Ninth {AAAI} Symposium
  on Educational Advances in Artificial Intelligence, {EAAI} 2019, Honolulu,
  Hawaii, USA, January 27 - February 1, 2019.},  742--749.

\bibitem[\protect\citeauthoryear{Welling and Teh}{2011}]{welling2011bayesian}
Welling, M., and Teh, Y.~W.
\newblock 2011.
\newblock Bayesian learning via stochastic gradient langevin dynamics.
\newblock In {\em Proceedings of the 28th international conference on machine
  learning (ICML-11)},  681--688.

\end{thebibliography}
